\documentclass[a4paper, 10pt, conference]{ieeeconf}      
\usepackage{FG2020}

\FGfinalcopy 

\IEEEoverridecommandlockouts                              
\def\FGPaperID{****} 

\title{\LARGE \bf
Spotting Macro- and Micro-expression Intervals in Long Video Sequences
}

\author{\parbox{16cm}{\centering
    {\large Ying He$^1$, Su-Jing Wang$^{1,2,*}$, Jingting Li$^1$ and Moi Hoon Yap$^3$}\\
    {\normalsize
    $^1$ Key Laboratory of Behavior Sciences, Institute of Psychology, Chinese Academy of Sciences, Beijing, China\\
    $^2$ Department of Psychology, University of the Chinese Academy of Sciences, Beijing, China\\
    $^3$ Department of Computing and Mathematics, Manchester Metropolitan University, Manchester, UK}}
    \thanks{This paper is supported by grants from the National Natural Science Foundation of China (U19B2032, 61772511) and the Royal Society UK (IF160006).}
    \thanks{$^*$Corresponding E-mail: wangsujing@psych.ac.cn.}
}

\usepackage{graphicx}
\usepackage{subfigure}
\usepackage{amsmath}
\usepackage{url}
\usepackage[colorlinks,urlcolor=magenta,linkcolor=black,anchorcolor=black,citecolor=black]{hyperref}

\begin{document}

\ifFGfinal
\thispagestyle{empty}
\pagestyle{empty}
\else
\author{Anonymous FG2020 submission\\ Paper ID \FGPaperID \\}
\pagestyle{plain}
\fi
\maketitle

\begin{abstract}
This paper presents baseline results for the Third Facial Micro-Expression Grand Challenge (MEGC 2020). Both macro- and micro-expression intervals in CAS(ME)\(^2\) and SAMM Long Videos are spotted by employing the method of Main Directional Maximal Difference Analysis (MDMD). The MDMD method uses the magnitude maximal difference in the main direction of optical flow features to spot facial movements. The single-frame prediction results of the original MDMD method are post-processed into reasonable video intervals. The metric F1-scores of baseline results are evaluated: for CAS(ME)\(^2\), the F1-scores are 0.1196 and 0.0082 for macro- and micro-expressions respectively, and the overall F1-score is 0.0376; for SAMM Long Videos, the F1-scores are 0.0629 and 0.0364 for macro- and micro-expressions respectively, and the overall F1-score is 0.0445. The baseline project codes are publicly available at \url{https://github.com/HeyingGithub/Baseline-project-for-MEGC2020_spotting}.
\end{abstract}

\section{Introduction}
Facial expressions are important non-verbal cues that convey emotions. Macro-expressions are the common facial expressions in our daily life, which are the types we usually know. There is a special type of expressions called ``micro-expressions" that were first found by Haggard and Isaacs~\cite{firstFind}. Micro-expressions (MEs) are involuntary facial movements occurring spontaneously when a person attempts to conceal the experiencing emotion in a high-stakes environment. The duration of MEs is very short. The general duration is less than 500 milliseconds (ms)~\cite{time1,time2}. The close connection between MEs and deception makes the relevant research have great significance on many applications such as medical care~\cite{appCare} and law enforcement~\cite{appLaw}.

Spotting expressions is to find the moment when expressions occur in the whole video sequences. In the past decades, several explorations for spotting MEs have been done: Early work studied on posed MEs~\cite{posedMethod1, posedMethod2, posedMethod3, posedMethod4, posedMethodStrain}. After several spontaneous ME datasets were established~\cite{SMIC, CASMEI, CASMEII, CASME2, SAMM}, later work explored ME spotting methods on the spontaneous MEs. Some methods\cite{X2, MESTowards, MDMD} designed a hand-crafted value to measure the difference between frames, and then used the value to spot ME frames. Some methods spotted ME frames based on a neural network~\cite{MES_DL_dense, SMEConvNet}. In the Second Micro-Expression Spotting Challenge (MEGC 2019)~\cite{MEGC2019}, a method for spotting ME intervals in long videos were explored~\cite{MESLongvideos}.

However, MEs are often accompanied by macro-expressions, and both of the two types of expressions are valuable for affect analysis. Therefore, developing methods to spot both macro- and micro-expressions is the main theme of the Third Facial Micro-Expression Grand Challenge (MEGC 2020).

In this paper, we provide the baseline method and results for MEGC 2020, spotting macro- and micro-expression intervals in long video sequences from the dataset CAS(ME)\(^2\) and SAMM Long Videos. The main method is Main Directional Maximal Difference Analysis (MDMD)~\cite{MDMD}. The original MDMD only predicts whether a frame belongs to facial movements. To obtain target intervals, the adjacent frames consistently predicted to be macro- or micro-expressions form an interval, and the intervals that are too long or too short are removed. Parameters are adjusted to specific expression types for specific datasets. The performance metric, F1-scores, is used for the evaluation on the two long video datasets.

The rest of paper is organized as follows: Section~\ref{sec:challenge} presents the challenge details. Section~\ref{sec:method}  introduces the the baseline method. Section~\ref{sec:result} shows the the experiment results. Section~\ref{sec:conclusion} concludes the paper.

\section{Challenge Details}\label{sec:challenge}
This section introduces the benchmark datasets and the performance metrics.
\subsection{Datasets}
\textbf{CAS(ME)\(^2\)} \cite{CASME2}: In the part A of CAS(ME)$^2$ database, there are 22 subjects and 98 long videos. The facial movements are classified as macro- and micro-expressions. The video samples may contain multiple macro or micro facial expressions. The onset, apex, offset indexes for these expressions are given in the excel file. In addition, eye blinks are labeled with the onset and offset time. The offset is labeled with 0 when a macro-expression continues and doesn't end. When it happens, we make the true interval be [onset, apex] (i.e. the offset equals the apex).

\textbf{SAMM Long Videos} \cite{yap2019samm} : The original SAMM dataset \cite{SAMM} contains 159 micro-expressions, which was used for the past two micro-expression recognition challenge \cite{MEGC2018, MEGC2019}. Recently, the authors \cite{yap2019samm} released the SAMM Long Videos dataset, which consists of 147 long videos. There are 343 macro-movements and 159 micro-movements in the long videos. The indexes of onset, apex and offset frames of micro- and macro-movements are outlined in the ground truth excel file.

More detailed and comparative information of these two datasets is presented in Table \ref{tab:dt_info}.

\begin{table}[h]
\caption{A Comparison between CAS(ME)$^2$ and SAMM Long Videos.}
\centering
\begin{tabular}{|l|l|l|}
\hline
Dataset & \textbf{CAS(ME)$^2$} & \textbf{SAMM Long Videos} \\ \hline
Participants & 22 & 32 \\ \hline
Video samples & 98 & 147 \\ \hline
Macro-expressions & 300 & 343 \\ \hline
Micro-expressions & 57 & 159 \\ \hline
Resolution & 640\(\times\)480 & 2040\(\times\)1088 \\ \hline
FPS & 30 & 200 \\ \hline
\end{tabular}\label{tab:dt_info}
\end{table}

\subsection{Performance metrics}
In order to avoid the inaccuracy caused by annotation, we propose to evaluate the spotting result per interval in MEGC 2020.
\subsection*{1. True positive in one video definition}
The true positive (TP) per interval in one video is first defined based on the intersection between the spotted interval and the ground-truth interval. The spotted interval $W_{spotted}$ is considered as TP if it fits the following condition:
\begin{equation}
   \frac{W_{spotted}\cap W_{groundTruth}}{W_{spotted}\cup W_{groundTruth}} \geq k
\end{equation}
where $k$ is set to 0.5, and $W_{groundTruth}$ represents the ground truth of the macro- or micro-expression interval (onset-offset). If the condition is not fulfilled, the spotted interval is regarded as a false positive (FP).

\subsection*{2. Result evaluation in one video}
Suppose there are $m$ ground truth intervals in the video, and $n$ intervals are spotted. According to the overlap evaluation, the TP amount in one video is counted as $a$ ($a\leq m$ and $a\leq n$), therefore FP = $n-a$, FN = $m-a$. The spotting performance in one video can be evaluated by following metrics:
\begin{equation}
    Recall = \frac{a}{m}, Precision = \frac{a}{n}
\end{equation}

\begin{equation}
    F-score = \frac{2TP}{2TP+FP+FN} = \frac{2a}{m+n}
\end{equation}
\par
Yet, the videos in real life have some complicated situations which influences the evaluation per single video:
\begin{itemize}
    \item There might be no macro- nor micro-expression in the test video. In this case, $m=0$, the denominator of recall would be zeros.

    \item
    If there is no spotted intervals in the video, the denominator of precision would be zeros since $n=0$.

    \item It is impossible to compare two spotting methods when both TP amounts are zero. The metric (recall, precision or F1-score) values both equal to zeros. However, the Method$_1$ outperforms Method$_2$, if Method$_1$ spots less intervals than Method$_2$.
\end{itemize}
\par
Thus, to avoid these situations, we propose for single video spotting result evaluation, we just note the amount of TP, FP and FN. Other metrics are not considered for one video.
\par
\subsection*{3. Evaluation for entire dataset}
Suppose in the entire dataset:

\begin{itemize}
    \item There are $V$ videos including $M_1$ macro-expressions (MaEs) sequences and $M_2$ micro-expression (MEs) sequences, where \[M_1 =\sum^V_{i=1} m_{1i} \text{ and } M_2 =\sum^V_{i=1} m_{2i};\]

    \item The method spot $N_1$ MaE intervals and $N_2$ ME intervals in total, where \[N_1 =\sum^V_{i=1} n_{1i} \text{ and } N_2 =\sum^V_{i=1} n_{2i};\]

    \item There are $A_1$ TPs for MaE and $A_2$ TPs for ME in total, where \[A_1 =\sum^V_{i=1} a_{1i} \text{ and } A_2 =\sum^V_{i=1} a_{2i}.\]
\end{itemize}
The dataset could be considered as one long video. The results are firstly evaluated for the MaE spotting and ME spotting separately. Then the overall result for macro- and micro spotting is evaluated. The \textit{recall} and \textit{precision} for entire dataset can be calculated by following formulas:
\begin{itemize}
    \item for macro-expression:
        \begin{equation}
            Recall_{MaE\_D} =\frac{A_1}{M_1}, Precision_{MaE\_D} = \frac{A_1}{N_1}
        \end{equation}

    \item for micro-expression:
    \begin{equation}
        Recall_{ME\_D} = \frac{A_2}{M_2}, Precision_{ME\_D} = \frac{A_2}{N_2}
    \end{equation}

\item for overall evaluation:
    \begin{equation}
        Recall_{D} = \frac{A_1+A_2}{M_1+M_2}, Precision_{D} = \frac{A_1+A_2}{N_1+N_2}
    \end{equation}
\end{itemize}
\par
Then, the values of \textit{F1-score} for all these three evaluations are obtained based on:
\begin{equation}
    F1-score = \frac{2\times (Recall \times Precision)} {Recall+Precision}
\end{equation}

The champion of the challenge will be the best score for overall results in spotting micro- and macro-expressions.

\section{Baseline Method}\label{sec:method}

\subsection{Preprocess}
Expression spotting focuses on facial regions. So we preprocess every video sample by cropping and resizing facial regions in all frames. For each video, we locate the rectangular box that exactly bounds the facial region in the first frame, and then all the frames of the video are cropped and resized according to the box located in the first frame. We locate the bounding box according to facial landmarks detected by the corresponding function in the "Dlib" toolkit \cite{King2009Dlib}, as we found that applying a face detecting algorithm directly cannot behave very well. The preprocess details are as follows.

\begin{figure*}[htbp] \centering
\subfigure[] { \label{fig:pre1}
\includegraphics[height=0.45\columnwidth]{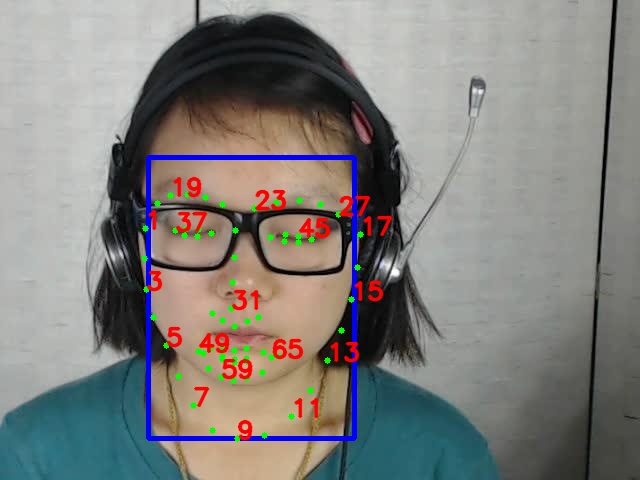}
}
\hspace{2mm}
\subfigure[] { \label{fig:pre2}
\includegraphics[height=0.45\columnwidth]{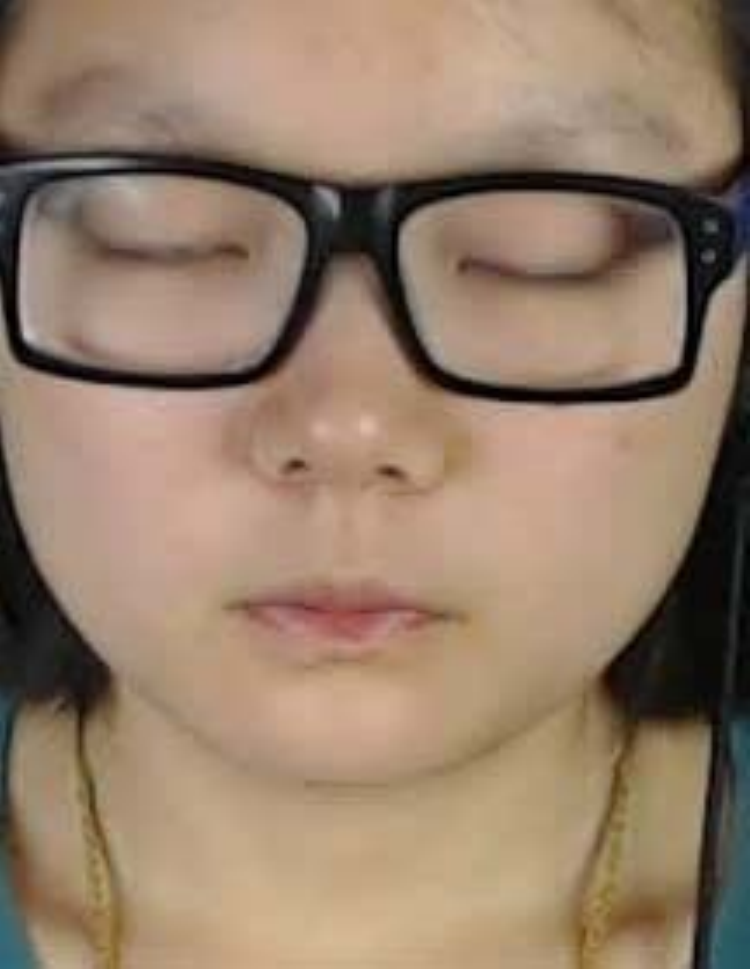}
}
\hspace{2mm}
\subfigure[] { \label{fig:pre3}
\includegraphics[height=0.45\columnwidth]{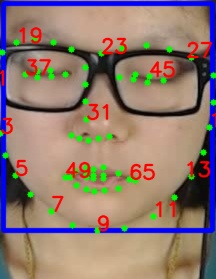}
}
\hspace{2mm}
\subfigure[] { \label{fig:pre4}
\includegraphics[height=0.45\columnwidth]{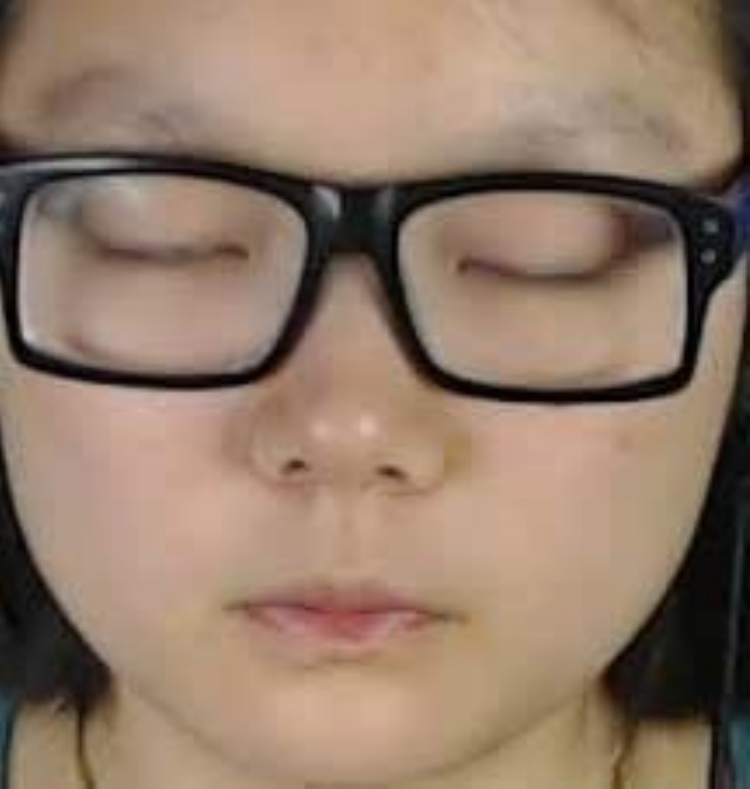}
}
\caption{Diagram of how we obtain facial regions in the preprocessing step: (a) detect facial landmarks and form the rectangular box \(\textbf{B}\); (b) the region in \(\textbf{B}\); (c) detect facial landmarks in the region in \(\textbf{B}\) and form the rectangular box \(\textbf{B}'\); (d) the region in \(\textbf{B}'\).}
\label{fig:preprocess}
\end{figure*}

Firstly, we use the landmark detecting function in the "Dlib" toolkit to obtain 68 facial landmarks on the face in the first frame of the video, as illustrated in Fig. \ref{fig:pre1} -- the first frame of s23\_0102 in CAS(ME)\(^2\). The landmarks are marked as \(L_1,L_2,\cdots,L_{68}\) in the sequence of the list returned by the landmark detection function in "Dlib", and the corresponding coordinates are marked as \((x_1,y_1), (x_2,y_2), \cdots, (x_{68},y_{68})\). The coordinate system is consistent with the one in the OpenCV toolkit~\cite{opencv_library}, i.e. x-axis means the horizontal direction from left to right, and y-axis means the vertical direction from top to bottom. The green dots in Fig.  \ref{fig:pre1} are the landmarks, and some of the serial numbers are marked by the red text.

Secondly, in order to form a rectangular box that bounds the facial region exactly, the leftmost, rightmost, topmost and bottommost landmarks are marked as \(L_l, L_r, L_t, L_b \) with coordinates \((x_l,y_l), (x_r,y_r),(x_t,y_t), (x_b,y_b) \), respectively. Rather than forming the box directly according to \(L_l, L_r, L_t, L_b \), we form two points: \(A(x_l,y_t-(y_{37}-y_{19})),B(x_r,y_b)\) to obtain the box \(\textbf{B}\) with \(A\) as the upper left corner and \(B\) as the lower right corner. The coordinate \(y_t-(y_{37}-y_{19})\) means that the upper edge of the box is moved up a relative distance to maintain more regions around eyebrows. In Fig. \ref{fig:pre1}, the box \(\textbf{B}\) is illustrated by the blue rectangular.

Thirdly, as shown by Fig. \ref{fig:pre2}, which is the region in \(\textbf{B}\), we found there are redundant regions in the bottom for several subjects in the two datasets because of the inaccuracy of landmark detection, and so, we detect landmarks again on the region of the first frame in \(\textbf{B}\) for cropping faces more precisely. It is shown in the Fig. \ref{fig:pre3}. Then, we get a new bottommost landmark \(L'_b(x'_b,y'_b)\). \(B\) is updated to \(B'(x_r,y_{min})\), where \(y_{min}\) is the smaller one of \(y_b\) and \(y'_b\). Then a new rectangular box \(\textbf{B}'\) is formed with \(A\) as the upper left corner and \(B'\) as the lower right corner. In Fig. \ref{fig:pre3}, the box \(\textbf{B}'\) is illustrated by the blue rectangular. And the region of the first frame in \(\textbf{B}'\) is illustrated in Fig. \ref{fig:pre4}, in which we can find that the facial region is located better.

Finally, after obtaining the box \(\textbf{B}'\), we crop all the frames of the video in the rectangular box \(\textbf{B}'\), and thus get the facial regions. The cropped regions are then resized to the size of \(227\times227\).

\subsection{MDMD}
The method of Main Directional Maximal Difference Analysis (MDMD) is proposed in the literature \cite{MDMD}. The main idea is that: when an expression happens, the face will experience a process of producing an expression and returning to a neutral face. The main movement directions will be opposite in the process. By analyzing it, expressions can be spotted. Here we review the MDMD method.

Given a video with \(n\) frames, the current frame is denoted as \(F_i\). \(F_{i-k}\) is the \(k\)-th frame before the \(F_i\), and \(F_{i+k}\) is the \(k\)-th frame after the \(F_i\). The robust local optical flow (RLOF) \cite{Senst2012Robust} between the \(F_{i-k}\) frame (Head Frame) and the \(F_i\) frame (Current Frame) is computed. We denote the optical flow by \((u^{HC}, v^{HC})\). For convenience, \((u^{HC}, v^{HC})\) means the displacement of any point. Similarly, the optical flow between the \(F_{i-k}\) frame (Head Frame) and the \(F_{i+k}\) frame (Tail Frame) is denoted by \((u^{HT}, v^{HT})\). Then, \((u^{HC}, v^{HC})\) and \((u^{HT}, v^{HT})\) are converted from Euclidean coordinates to polar coordinates \((\rho^{HC}, \theta^{HC})\) and \((\rho^{HT}, \theta^{HT})\), where \(\rho\) and \(\theta\) represent, respectively, the magnitude and direction.

\begin{figure}[htbp] \centering
\includegraphics[width=0.75\columnwidth]{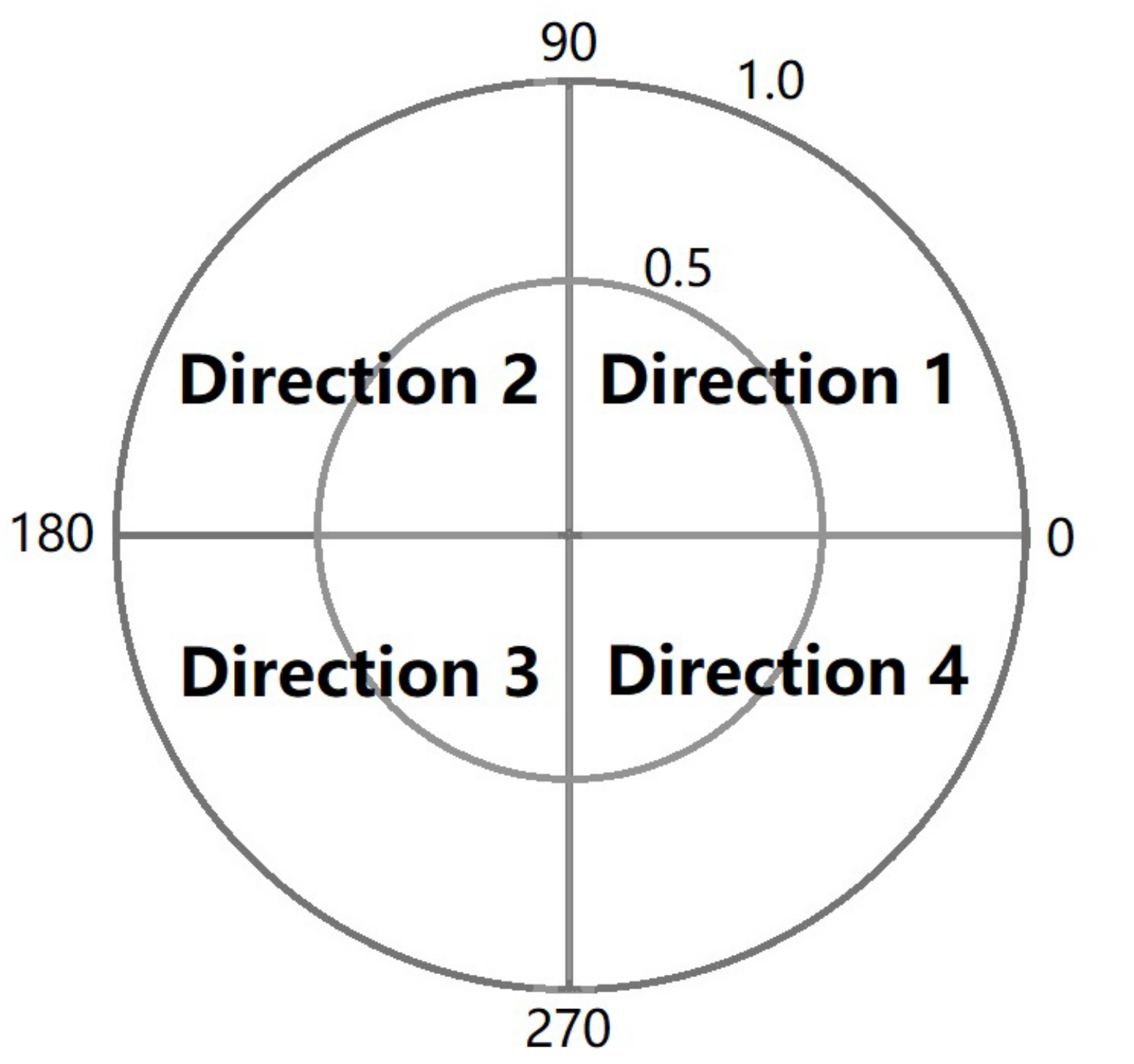}
\caption{Four directions in the polar coordinates.}
\label{fig:directions}
\end{figure}

Based on the directions \(\{\theta^{HC}\}\), all the optical flow vectors \(\{(\rho^{HC}, \theta^{HC})\}\) are divided into \(a\) directions. Fig. \ref{fig:directions} illustrates the condition when \(a=4\). The \emph{Main Direction} \(\Theta\) is the direction that has the largest number of optical flow vectors among the \(a\) directions. The main directional optical vector \((\rho^{HC}_M , \theta^{HC}_M )\) is the optical flow vector \((\rho^{HC}, \theta^{HC})\) that falls in the \emph{Main Direction} \(\Theta\).

\begin{equation}
\{(\rho^{HC}_M , \theta^{HC}_M )\}=\{(\rho^{HC} , \theta^{HC}) | \theta^{HC} \in \Theta\}
\label{eq:MDMD12}
\end{equation}
The optical flow vector corresponding to \((\rho^{HC}_M , \theta^{HC}_M )\) between \(F_{i-k}\) frame and \(F_{i+k}\) is denoted as \((\rho^{HT}_M , \theta^{HT}_M )\).

\begin{equation}
\begin{aligned}
\{(\rho^{HT}_M , \theta^{HT}_M )\}=&\{(\rho^{HT}, \theta^{HT}) | (\rho^{HT}, \theta^{HT}) \text{ and } (\rho^{HC}_M , \theta^{HC}_M ) \\
&\text{ are two different vectors of the same} \\
&\text{ point in } F_{i-k} \}
\label{eq:MDMD13}
\end{aligned}
\end{equation}
After the differences \(\rho^{HC}_M-\rho^{HT}_M\) are sorted in a descending order, the maximal difference \(d^i\) is defined as the mean difference value of the first 1/3 of the differences \(\rho^{HC}_M-\rho^{HT}_M\) to characterize the frame \(F_i\) as in the formula:

\begin{equation}
d=\frac{3}{g} \sum \max_{\frac{g}{3}} \{\rho^{HC}_M-\rho^{HT}_M\}
\label{eq:MDMD14}
\end{equation}
where \(g =|\{(\rho^{HC}_M-\rho^{HT}_M)\}|\) is the number of elements in the subset \(\{(\rho^{HC}_M-\rho^{HT}_M)\}\), and \(\max_mS\) denotes a set comprised of the first \(m\) maximal elements in the subset \(S\).

\begin{figure}[htbp] \centering
\includegraphics[width=0.55\columnwidth]{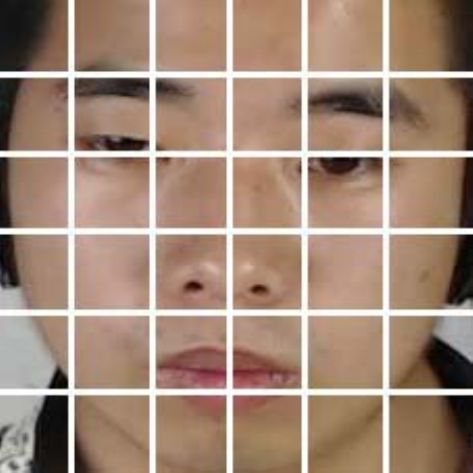}
\caption{An example of facial \(6\times6\) block structure.}
\label{fig:block}
\end{figure}

Since our method is a block-based analysis, the cropped facial region of each frame is divided into \(b{\times}b\) blocks, as shown in Fig. \ref{fig:block}. And we calculate the maximal difference \(d^i_j (j = 1, 2, \cdots, b^2)\) for each block in the \(F_i\) frame. For frame \(F_i\), there are \(b^2\) maximal differences \(d^i_j\) due to the \(b{\times}b\) block structure. Then, we arrange the \(b^2\) maximal differences \(d^i_j\) in a descending order where \(\bar d^i\) is the first 1/3 of the maximal differences and characterizes the frame \(F_i\) feature:

\begin{equation}
\bar d^i = \frac{1}{s}\sum \max_s\{d^i_j\},
\label{eq:MDMD15}
\end{equation}
where \(j = 1,2,\cdots,b^2\).

\begin{figure*}[htbp] \centering
\subfigure[] { \label{fig:situation1}
\includegraphics[width=1.5\columnwidth,height=0.53\columnwidth]{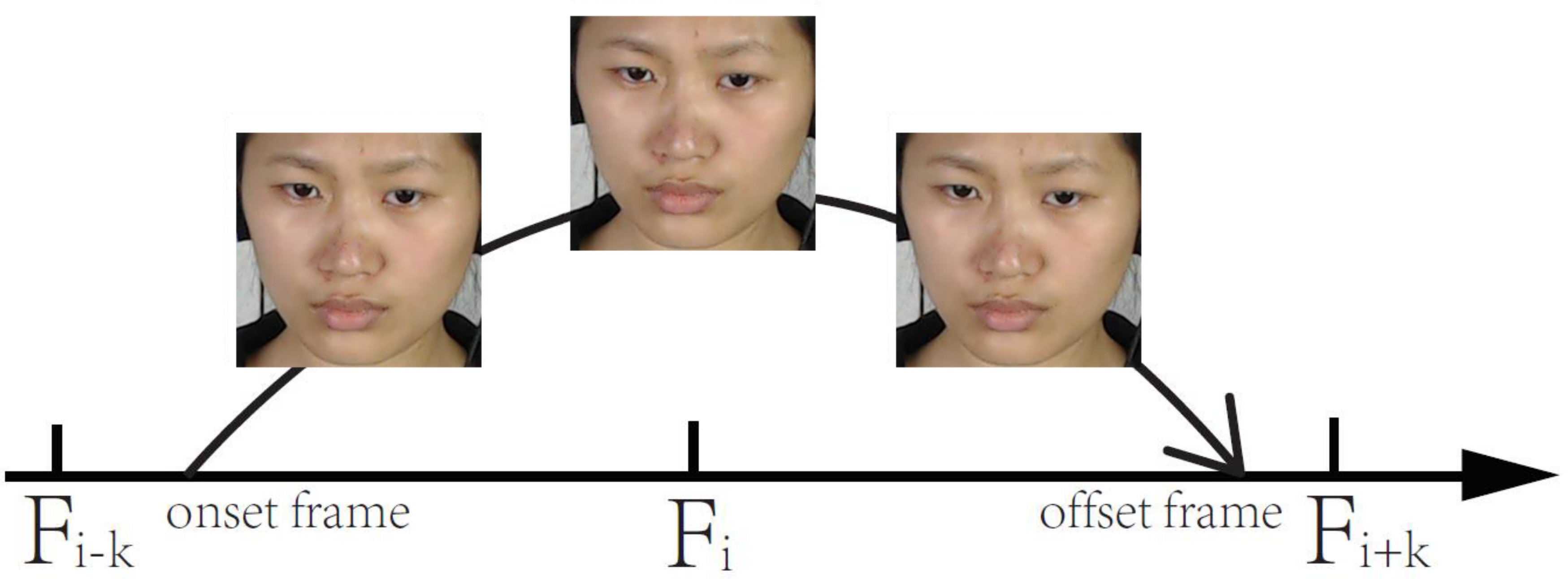}
}

\subfigure[] { \label{fig:situation2}
\includegraphics[width=1.5\columnwidth,height=0.53\columnwidth]{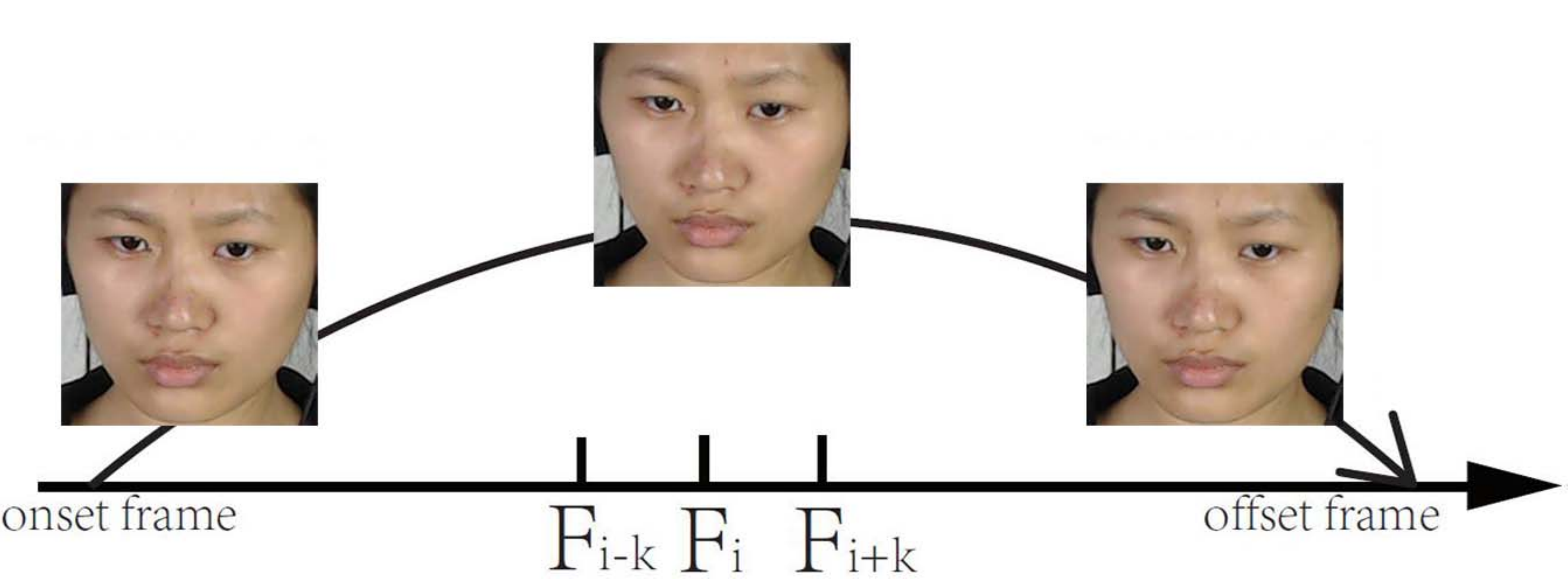}
}
\caption{Two situations: (a) An emotional expression starting at the onset frame between \(F_{i-k}\) and \(F_i\) is repressed at the offset frame between \(F_i\) and \(F_{i+k}\) and recovers a neutral expression at \(F_{i+k}\); (b) An emotional expression starting at the onset frame before \(F_{i-k}\) is repressed at the offset frame after \(F_{i+k}\).}
\label{fig:situations}
\end{figure*}

If a person maintains a neutral expression at \(F_{i-k}\), her/his emotional expression, such as disgust, starts at the onset frame between \(F_{i-k}\) and \(F_i\), and is repressed at the offset frame between \(F_i\) and \(F_{i+k}\), and then the facial expression recovers a neutral expression at \(F_{i+k}\), which is presented in Fig. \ref{fig:situation1}. In this circumstance, the movement between \(F_i\) and \(F_{i-k}\) is more intense than the movement between \(F_{i+k}\) and \(F_{i-k}\) because the expression is neutral at both \(F_{i+k}\) and \(F_{i-k}\). Therefore, the \(\bar d^i\) value will be large. Another situation is that a person maintains a neutral expression from \(F_{i-k}\) to \(F_{i+k}\). The movement between \(F_i\) and \(F_{i-k}\) is similar to the movement between \(F_{i+k}\) and \(F_{i-k}\); thus, the \(\bar d^i\) value will be small. In a long video, sometimes an emotional expression starts at the onset frame before \(F_{i-k}\) and is repressed at the offset frame after \(F_{i+k}\), which is presented in Fig. \ref{fig:situation2}. In this case, the \(\bar d^i\) value will also be small if \(k\) is set to be a small value. However, \(k\) cannot be set as a large value because this would influence the accuracy of the computing optical flow.

We employ a relative difference vector for eliminating the background noise, which is computed by:

\begin{equation}
r^i = \overline d^i - \frac{1}{2}\left(\overline d^{i-k+1} + \overline d^{i+k-1} \right),
\label{eq:MDMD16}
\end{equation}
where \(i=k+1,k+2,\cdots,n-k\).

Therefore, the frame \(F_i\) is characterized by \(r^i\). A threshold is used to obtain the frames that have peaks representing the facial movements in a video:

\begin{equation}
threshold = r_{mean} + p \times (r_{max} - r_{mean})
\label{eq:thresh}
\end{equation}
where \[r_{mean} = \frac{1}{n-2k}\sum _{n-k}^{i=k+1}r^i\] and \[r_{max} = \max_{n-k}^{i=k+1}r^i. \] \(p\) is a variable parameter in the range \([0,1]\). The frames with \(r^i\) larger than the \(threshold\) are the frames where expressions appear.

\subsection{Parameter settings and post process}
In the literature \cite{MDMD}, several parameter combinations are explored to spot micro-expressions on the CAS(ME)\(^2\) dataset. For spotting both macro- and micro-expressions on the two datasets for MEGC 2020, i.e. CAS(ME)\(^2\) and SAMM Long Videos, we select the best combination of blocks and directions explored in \cite{MDMD}, and we set other parameters according to the video FPSs of the two datasets. Moreover, since the original MDMD only predicts whether a frame belongs to facial movements, a post process is added in order to output target intervals required by MEGC 2020. The details are as follows.

The number of blocks is set to \(6\times6\) and the number of directions \(a\) is set to 4. In CAS(ME)\(^2\), the \(k\) is set to 12 for micro-expressions, and 39 for macro-expressions; in SAMM Long Videos, the \(k\) is set to 80 for micro-expressions, and 260 for macro-expressions. Concerning the threshold, \(p\) varies from 0.01 to 0.99 with a step-size of 0.01. And the final results are reported under the setting of \(p=0.01\). The original MDMD only predicts whether a frame belongs to facial movements. To output target intervals, the adjacent frames consistently predicted to be macro- or micro-expressions form an interval, and the intervals that are too long or too short are removed. The number of micro-expression frames is limited between 7 and 16 for the CAS(ME)\(^2\) dataset, and between 47 and 105 for the SAMM Long Videos dataset. The number of macro-expression frames is defined as larger than 16 for the CAS(ME)\(^2\) dataset, and larger than 105 for the SAMM Long Videos dataset.

\section{Results and Discussion}\label{sec:result}
\begin{table*}[htbp]
\caption{Baseline results in CAS(ME)\(^2\) and SAMM Long Videos with \(p\) varying from 0.01 to 0.20 with a step-size of 0.01.}
    \centering
    \begin{tabular}{|c|cc|cc|cc|cc|}
        \hline
        Dataset & \multicolumn{4}{|c|}{\textbf{CAS(ME)\(^2\)}} & \multicolumn{4}{|c|}{\textbf{SAMM Long Videos}} \\
        \hline
        Expression & \multicolumn{2}{|c|}{macro-expression} & \multicolumn{2}{|c|}{micro-expression} & \multicolumn{2}{|c|}{macro-expression} & \multicolumn{2}{|c|}{micro-expression} \\
        \hline
        \(p\) (\%) &TP &F1-score &TP &F1-score &TP &F1-score &TP &F1-score \\ \hline
1 & \textbf{109} & 0.1196 & \textbf{21} & 0.0082 & \textbf{22} & \textbf{0.0629} & \textbf{29} & \textbf{0.0364 }\\
2 & 107 & 0.1408 & 18 & 0.0093 & 20 & 0.0627 & 25 & 0.0356 \\
3 & 96 & 0.1455 & 15 & 0.0100  & 18 & 0.0627 & 19 & 0.0309 \\
4 & 92 & 0.1573 & 14 & 0.0115 & 16 & 0.0588 & 17 & 0.0306 \\
5 & 91 & 0.1738 & 12 & 0.0121 & 16 & 0.0626 & 14 & 0.0282 \\
6 & 88 & 0.1857 & 10 & 0.0120  & 14 & 0.0574  & 11 & 0.0245 \\
7 & 81 & 0.1879 & 10 & 0.0142 & 12 & 0.0510  & 11 & 0.0266 \\
8 & 74 & 0.1876 & 8 & 0.0131 & 10 & 0.0443 & 9 & 0.0239 \\
9 & 73 & 0.1984 & 8 & 0.0155 & 9 & 0.0407 & 7 & 0.0201 \\
10 & 68 & 0.1954 & 8 & 0.0176 & 8 & 0.0371 & 7 & 0.0214 \\
11 & 61 & 0.1863 & 6 & 0.0150  & 8 & 0.0378 & 7 & 0.0228 \\
12 & 61 & \textbf{0.2013} & 6 & 0.0173 & 8 & 0.0382 & 7 & 0.0245 \\
13 & 57 & 0.1949 & 6 & 0.0190  & 7 & 0.0337 & 6 & 0.0219 \\
14 & 56 & 0.2007 & 6 & 0.0214 & 7 & 0.0340  & 6 & 0.0227 \\
15 & 50 & 0.1859 & 5 & 0.0197 & 6 & 0.0299 & 5 & 0.0200  \\
16 & 50 & 0.1927 & 5 & 0.0214 & 6 & 0.0301 & 5 & 0.0210  \\
17 & 48 & 0.1886 & 5 & 0.0236 & 6 & 0.0304 & 5 & 0.0222 \\
18 & 46 & 0.1855 & 5 & 0.0253 & 6 & 0.0305 & 4 & 0.0183 \\
19 & 43 & 0.1795 & 5 & \textbf{0.0275} & 6 & 0.0310  & 3 & 0.0146 \\
20 & 42 & 0.1783 & 3 & 0.0179 & 6 & 0.0313 & 3 & 0.0152 \\ \hline
    \end{tabular}
    \label{tab:p_vary}
\end{table*}

\begin{table*}[htbp]
\caption{Baseline results for macro- and micro-spotting (\(p=0.01\)) in CAS(ME)\(^2\) and SAMM Long Videos.  }
    \centering
    \begin{tabular}{|c|c|c|c|c|c|c|c|}
        \hline
        Dataset & \multicolumn{3}{|c|}{\textbf{CAS(ME)\(^2\)}}& \multicolumn{3}{|c|}{\textbf{SAMM Long Videos}} \\ \hline
        Expression &macro-expression &micro-expression &overall result &macro-expression &micro-expression &overall result \\ \hline
        Total number &300 &57 &357 &343 &159 &502 \\ \hline
        TP &109 &21 &130 &22 &29 &51 \\ \hline
        FP &1414 &5014 &6428 &334 &1407 &1741 \\ \hline
        FN &191 &36 &227 &321 &130 &451 \\ \hline
        Precision &0.0716 &0.0042 &0.0198 &0.0618 &0.0202 &0.0285 \\ \hline
        Recall &0.3633 &0.3684 &0.3641 &0.0641 &0.1824 &0.1016 \\ \hline
        F1-score &0.1196 &0.0082 &0.0376 &0.0629 &0.0364 &0.0445 \\ \hline
    \end{tabular}
    \label{tab:result}
\end{table*}

For the parameter \(p\), we have studied the evaluation results by varying \(p\) from 0.01 to 0.99 with a step-size of 0.01, and the 20 results from 0.01 to 0.20 are shown in Table~\ref{tab:p_vary}. In Table~\ref{tab:p_vary}, we list the information of TPs and F1-scores for macro- and micro-expression spotting respectively. We observe that, for both types of expressions in the two datasets, the number of TP is decreasing with the increase of \(p\). Regarding the F1-score, it also shows a decreasing trend in SAMM Long Videos. Yet, in CAS(ME)\(^2\), the F1-score increases at first and then begins to decrease. The initial increase of the F1-score in CAS(ME)\(^2\) is mainly because the number of the total predicted intervals (\(n\)) becomes smaller with the increase of \(p\), making the precision (\(a/n\)) increase.

Since the amount of TP is an important metric for the spotting result evaluation, we select the results under the condition of \(p=0.01\) as the final baseline results. The details of the final baseline results for spotting macro- and micro-expressions are shown in Table \ref{tab:result}. For CAS(ME)\(^2\), the F1-scores are 0.1196 and 0.0082 for macro- and micro-expressions respectively, and 0.0376 for the overall result. For SAMM Long Videos, the F1-scores are 0.0629 and 0.0364 for macro- and micro-expressions respectively, and 0.0445 for the overall result. More details about the number of true labels, TP, FP, FN, precision, recall, and F1-score for various situations are shown in Table~\ref{tab:result}.

\section{Conclusions}\label{sec:conclusion}
This paper addresses the challenge in spotting macro- and micro-expressions in long video sequences, and provides the baseline method and results for the Third Facial Micro-Expression Spotting Challenge (MEGC 2020). The Main Directional Maximal Difference Analysis (MDMD) \cite{MDMD} is employed as the baseline method, and the parameter settings are adjusted to CAS(ME)\(^2\) and SAMM Long Videos for the spotting challenge in MEGC 2020. Slight modification is done to predict more reasonable intervals on the post-processing of results. Experiments were done and the predicted results were evaluated using the metrics in MEGC 2020. The results have shown that the MDMD method can produce reasonable performance, but there is still a huge challenge to reduce the amount of FPs.

\section{ACKNOWLEDGMENTS}
The authors gratefully acknowledge the contribution of reviewers' comments, which greatly improve the quality of this paper.

\bibliographystyle{ieee}
\bibliography{baseline_MEGC2020}

\end{document}